\documentclass{article}

\usepackage[final]{neurips_2021}

\usepackage[utf8]{inputenc} 
\usepackage[T1]{fontenc}    
\usepackage[colorlinks = true,
            linkcolor = blue,
            urlcolor  = blue,
            citecolor = red,
            anchorcolor = blue]{hyperref}
\usepackage{url}            
\usepackage{booktabs}       
\usepackage{amsfonts}       
\usepackage{amssymb}
\usepackage{nicefrac}       
\usepackage{microtype}      
\usepackage{xcolor}         
\usepackage{lipsum}
\usepackage{graphicx}
\usepackage{natbib}
\usepackage{mwe}
\usepackage{pifont}
\usepackage{caption}
\usepackage[ruled]{algorithm2e}
\usepackage{multirow}

\let\emptyset\varnothing

\SetKwComment{Comment}{/* }{ */}

\newcommand{\cmark}{\ding{51}}%
\newcommand{\xmark}{\ding{55}}%
\newcommand{\dataset}[1]{%
  \def\param{#1}%
  \def\raw{raw}%
  \def\cleaned{cleaned}%
  \def\hc{hc}%
  \def\lb{lb}%
  \def\synfirst{synfirst}%
  \def\synsecond{synsecond}%
  \def\pool{pool}%
  \ifx\raw\param$\mathit{Base_0}$\,\else
  \ifx\cleaned\param$\mathit{Base_1}$\,\else
  \ifx\hc\param$\mathit{Handcrafted}$\,\else
  \ifx\lb\param$\mathit{Label\_book}$\,\else
 \ifx\synfirst\param$\mathit{Synthetic_1}$\,\else
 \ifx\synsecond\param$\mathit{Synthetic_2}$\,\else
 \ifx\pool\param$\mathit{Pool}$\,\else
  $\mathit{Dataset_{#1}}$%
  \fi\fi\fi\fi\fi\fi\fi
}

\title{Increasing Data Diversity with Iterative Sampling to Improve Performance}

\author{%
  Devrim Cavusoglu\\
  \And
  Oğulcan Eryuksel \\
  \And
  Sinan Altinuc \\
  \AND
  \vspace{-0.5cm}\\
  OBSS AI\\
  \texttt{\{devrim.cavusoglu, ogulcan.eryuksel, sinan.altinuc\}@obss.com.tr} \\
} 

\begin{document}

\maketitle

\begin{abstract}
  As a part of the Data-Centric AI Competition, we propose a data-centric approach to improve the diversity of the training samples by iterative sampling. The method itself relies strongly on the fidelity of augmented samples and the diversity of the augmentation methods. Moreover, we improve the performance further by introducing more samples for the difficult classes especially providing closer samples to edge cases potentially those the model at hand misclassifies.
\end{abstract}

\section{Introduction}

This paper serves as a summary documentation of methods for experiments studied under the Data-Centric AI competition \cite{dccomp2021}. The competition provides an MNIST \cite{lecun1998gradient} style dataset with Roman numerals and expects the competitors to increase the performance by only manipulating data and not the code or the model to emphasize the importance of data in machine learning models and enforce a data-centric approach. The competition holds a fixed model and training setup for all submissions.

\section{Experimental Setup and Tools}

The official training script\footnote{The official training script can be accessed \href{https://worksheets.codalab.org/bundles/0x57030e4a2d034af4b8efa38df4ff6af6}{here}.} is used in all training phases of the experiments unless stated otherwise. The model is part of the ResNet50 \cite{he2016deep} as defined in the official training script. Without changing the training setup in the file, we added some utilities and helpers such as training accuracy/loss plots, classification reports and export of prediction outputs to ease the interpretation and exploration of the outputs.

The training loss/accuracy plot is added to track how well we develop and iterate the data set. To see how well the model performs on different classes, we added a classification report \cite{scikit-learn} to better understand and aid the data. To better discriminate the misclassified samples visually, we used FiftyOne package \cite{moore2020fiftyone}. FiftyOne allowed us to go through the samples easily for which parts we wanted to investigate. We also utilize FiftyOne to compute similarities of embeddings for iterative sampling, which is discussed under \autoref{sec:is}.

\section{Evolution of Dataset(s) \& Applied Methods}

The information for all datasets used, manipulated, or generated during experiments are listed on \autoref{tab:data-info}. Our validation set is constructed from the original validation split of \dataset{raw} by adding augmented samples until each class has the same amount of samples, 100. We use \dataset{lb} given under the competition as "Label book" in the experiments and results.

\begin{table}
\centering
\resizebox{1.1\textwidth}{!}{%
\hspace{-1.5cm}
\begin{tabular}{ccccc}
    \hline
    Dataset name & Equal & Avg. size per class (train) & Size per class (val) & Description\\
    \hline
    \dataset{raw}  & \xmark & 207.2 & 81.6 & Original competition images \\
    \dataset{cleaned} & \cmark & 250 & 100 & Cleaned \dataset{raw} + Fixed per class size with DA \\
    \dataset{hc} & \cmark & 30 & - & Handwritten Roman Numerals by People\\
    \dataset{lb} & \xmark & 5.2 & - & Label book provided by the competition\\
    \dataset{synfirst} & \xmark & 78 & - & Synthetic Handwriting Samples with $Style\_Set_1$\\
    \dataset{synsecond} & \xmark & 124 & - & Synthetic Handwriting Samples with $Style\_Set_2$\\
    \multirow{2}{*}{\dataset{pool}} & \multirow{2}{*}{\cmark} & \multirow{2}{*}{6782.7} & \multirow{2}{*}{-} & Augmented image samples from the combination of \\
             &        &      &   & \dataset{raw} + \dataset{hc} + \dataset{synfirst} + \dataset{synsecond}\\
    \dataset{syn1} & \xmark & 328 & 100 & \dataset{cleaned} + \dataset{synfirst}\\
    \dataset{syn2} & \xmark & 452 & 100 & \dataset{syn1} + \dataset{syn2}\\
    \dataset{syn\_aug} & \cmark & 900 & 100 & \dataset{syn2} + Augmentations\\
    \dataset{iter} & \cmark & 900 & 100 & \dataset{syn\_aug} + Iterative Sampling\\
    \dataset{uneven} & \xmark & 900 & 100 & \dataset{iter} + Favoring difficult classes\\
    \hline
\end{tabular}%
}
\vspace{0.2cm}
\caption{Description of datasets used during the competition. Column "Equal" indicates if training sample sizes are equal for each classes or not, note that for validation split this is always true except \dataset{raw}. Only train splits are manipulated and validation (and test) splits are held fixed except \dataset{raw} which is raw dataset provided by the competition.}
\label{tab:data-info}
\end{table}

Naturally, the development process for enhancing the data (specifically training set since validation and test sets are held fixed in our experiments) goes sequentially cumulative. That is, until reaching the hard limit for the competition submission, which is 10,000 (9,000 for training) samples in total, we tried to introduce each method with a soft limit on top of the previous one. This procedure is held for all datasets generated except for the final method, which is changing class sizes so that difficult classes have more training samples by visual inspection.

\subsection{Cleaning the Raw Dataset}
\begin{figure}
    \centering
    \begin{minipage}{0.2\textwidth}
        \hspace{-0.75cm}\includegraphics[width=1.1\textwidth]{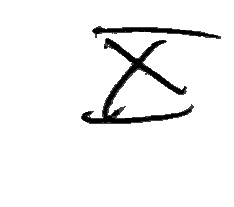}
        \caption*{\textbf{(a)} Case 1}
    \end{minipage}\hfill
    \begin{minipage}{0.2\textwidth}
        \hspace{-3.5cm}\includegraphics[width=1.1\textwidth]{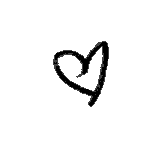}
        \caption*{\hspace{-6.5cm}\textbf{(b)} Case 2}
    \end{minipage}
    \begin{minipage}{0.2\textwidth}
        \hspace{-1.5cm}\includegraphics[width=1.2\textwidth]{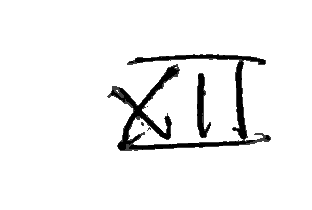}
        \caption*{\hspace{-1.5cm}\textbf{(c)} Case 3}
    \end{minipage}
    \caption{Three example cases for bad samples in \dataset{raw}. Sample \textbf{(a)} belongs to class "iv" whereas it actually belongs to class "x". Sample \textbf{(b)} is not a Roman numeral, but present under class "vii". Sample \textbf{(c)} belongs to class "vii" whereas it is distinguishably not a "vii".}
    \label{fig:bad-samples}
\end{figure}

To create the \dataset{cleaned} dataset, we removed samples that are non-numeral or too ambiguous between classes and corrected the labels for mislabelled examples. It can be observed in \autoref{fig:bad-samples} that some samples violate the consistency and validity of the dataset.

\subsection{Synthetic Handwriting Generation}
We used a synthetic handwriting generative model \cite{graves2013generating} to enrich the dataset diversity in the training split. We use the implementation of \cite{svasquez-handwriting}. One drawback for this model is the tendency to add more characters than intended.

\begin{figure}
    \centering
    \begin{minipage}{0.2\textwidth}
        \hspace{5cm}\includegraphics[width=1\textwidth]{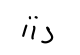}
    \end{minipage}\hfill
    \begin{minipage}{0.2\textwidth}
        \hspace{-3cm}\includegraphics[width=1\textwidth]{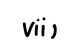}
    \end{minipage}
    \begin{minipage}{0.2\textwidth}
        \hspace{-0.5cm}\includegraphics[width=1\textwidth]{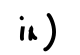}
    \end{minipage}
    \caption{Samples with discrepancy that have trailing extra components.}
    \label{fig:bad-samples-rnn}
\end{figure}

The generated samples from RNN are quite satisfying; however, some of the generated samples for some set of classes have discrepancies which can be seen in \autoref{fig:bad-samples-rnn}. Some of these samples with flaws are corrected by adding a suffix to text and applying traditional post-processing to remove that suffix from the generated image, and to discard all flawed generated samples, the generated samples are visually inspected after the post-processing.  

The implementation we use has a total of 13 styles and additional parameters such as $bias$ and $stroke\_width$. These parameters are either fixed to a certain value or set to a distribution where the distribution parameters are fixed by visually inspecting the generated samples. By using two different style sets, we generated \dataset{synfirst} and \dataset{synsecond} dataset. With these two synthetic datasets, we generated \dataset{syn1} and \dataset{syn2} by combining them with the cleaned dataset \dataset{cleaned}. 

\subsection{Data Augmentation}
\label{sec:sda}

In order to both populate and enhance the diversity of the dataset, we applied data augmentation on \dataset{syn2}. We selected some proper set of augmentations for these samples, which are grayscale and generally in small sizes (resolution). We used Augly package \cite{bitton2021augly} for our augmentation procedure. The set of augmentations we used are \textit{HFLip, VFlip, ShufflePixels, Pixelization, Rotation, Blur, RandomAspectRatio, Noise} from Augly. We also used barrel and barrel inverse distortions from ImageMagick \cite{imagemagick}.

Note that we used these augmentations as single or as a composition of several. For these augmentations, we manually and in an iterative way selected a certain set of parameters for each of them carefully not to make the image unrecognizable. The potential danger zone here is that after an augmentation, the resulting image may become unrecognizable or may belong to a different class (horizontal flip of "vi" and "iv"). With the specified augmentation list, we combine the augmented images with \dataset{syn2} so that augmented images fill the gap until maximum value per class, \textit{900} (maximum class size for equal class sizes). With this combination we get \dataset{syn\_aug}.

\subsection{Iterative Sampling with Augmentation}
\label{sec:is}

To increase diversity even more, we propose an algorithm that iteratively removes the similar samples and adds from a pool of augmented samples. We first created a pool of augmented images, \dataset{pool}, with the augmentation set mentioned in \autoref{sec:sda}. \dataset{pool} contains only augmented samples and not the base images. Besides the augmented image pool, we also need embeddings for distance calculation. To get embeddings, we used the official model, trained on \dataset{syn\_aug}, and we got the global average pooling outputs, $layer^{L-1}$ where $layer^L$ is the softmax layer. The embeddings for samples are vectors of $dim (1,256)$.

\begin{algorithm}
\caption{Iterative sampling algorithm.}
\label{alg:sampling}
\KwData{$D$ dataset, $P$ augmented\_data\_pool, $N \ge 0$ number of iterations,\\ 
\hspace{1cm}$max\_sizes$ maximum size per class, $metric$ distance metric}
\KwResult{$D'$ new dataset}
$C \gets [i, ii, iii, iv, v, vi, vii, viii, ix, x]$\;
\For{c in C}{
    $D_c \gets get\_class\_samples(D, class=c)$\;
    $E_c \gets get\_embeddings(D_c)$ \Comment*[r]{retrieve model embeddings}
    $P_c \gets get\_class\_samples(P, class=c)$\;
    $S \gets find\_duplicates(E_c, metric)$\;
    $n \gets 1$\;
    \While{$S \neq \emptyset$ \textbf{or} $n \le N$}{
        $remove(D_c, S)$  \Comment*[r]{remove $S$ from $D_c$}
        $m \gets max\_sizes_c - size(D_c)$\;
        $P_{sub} \gets select(P_c, n=m)$ \Comment*[r]{randomly get $m$ samples from $P_c$}
        $add(D_c, P_{sub})$ \Comment*[r]{add samples from $P_{sub}$ to $D_c$}
        $remove(P_{sub}, P_c)$ \Comment*[r]{remove $P_{sub}$ from $P_c$}
        $S \gets find\_duplicates(E_c, metric)$\;
        $n \gets n + 1$\;
    }
}
\end{algorithm}

Then, using \autoref{alg:sampling} we iteratively replaced the similar samples in \dataset{syn\_aug} with the samples randomly drawn from \dataset{pool}. To find similar samples, we used FiftyOne's $find\_duplicate$ function. In this experiment, we used Euclidean distance as a distance metric and $N=10$. With this algorithm, \dataset{iter} is generated. Realize that the achievement of this procedure does not solely rely on the similarity computation by embeddings but also assuring the diversity of augmentations.

\subsection{Favoring Difficult Classes}
\label{sec:uneven}

By observing the classification performance of the model for each class by f1 metric, we have seen that the classification performance for the classes "i", "v", and "x" are relatively higher. Thus, we have decided to imbalance the sample sizes towards more difficult classes as apparently "i", "v", and "x" are easier for the model to learn compared to other classes. Then, visually examining the misclassified examples through FiftyOne, we manually selected samples to the misclassified ones from \dataset{pool}. With this approach, we retrieve \dataset{uneven}.

\section{Conclusion}
We took an iterative approach using the following methods in succession: data cleaning, generating synthetic data using RNN based handwriting generation, data augmentation, iterative sampling, creating an imbalance of classes favoring difficult ones. Each step provided a marginal improvement. However, it should be noted that it would not be very accurate to directly compare these improvements to each other as it gets harder to improve the accuracy as the overall accuracy increases. The results of the datasets can be seen in \autoref{tab:results}. Results of  \dataset{syn1} and  \dataset{syn2} Indicate that synthetic handwriting data generation strategy improved the results significantly providing a total accuracy improvement of 9.1 points. Marginally \dataset{iter} provides the most significant gain. Therefore, the proposed method of iterative sampling utilizing model embeddings improves the diversity of the dataset. The manipulated dataset \dataset{uneven} boosted the performance further and attained the highest score suggesting that having uneven class sizes favoring difficult classes can work well. Note that synthetic data generation and augmentation improve accuracy by increasing the number of samples; whereas, iterative sampling and favoring difficult classes increase performance without changing the total sample size.

This paper shows an approach to diversifying the dataset through iterative sampling, and also boost the performance even more with making class sizes uneven such that difficult classes have more samples. The results suggest that the approaches proposed here works well with the dataset (MNIST style with Roman Numerals) provided by the competition.

\begin{table}
    \centering
    \resizebox{1\textwidth}{!}{\hspace{-0cm}\begin{tabular}{c|ccccc}
    Dataset &  Train set Acc. & Validation set Acc. & Test set Acc. & Marginal Gain & Cumulative Gain\\
    \hline
    \dataset{raw} & 99.66 & 67.53 & 59.62 & - & -\\
    \dataset{cleaned} & \textbf{1.00} & 72.30 & 61.54 & - & -\\
    \dataset{syn1} & 99.75 & 77.80 & 67.30 & 5.5 & 5.5\\
    \dataset{syn2} & 99.67 & 81.40 & 73.08 & 3.6 & 9.1\\
    \dataset{syn\_aug} & 99.72 & 85.00 & 71.15 & 3.6 & 12.7\\
    \dataset{iter} & 99.28 & 93.20 & \textbf{98.08} & \textbf{8.2} & 20.9\\
    \dataset{uneven} & 99.31 & \textbf{95.50} & \textbf{98.08} & 2.3 & \textbf{23.2}\\
    \hline
    \end{tabular}}
    \vspace{0.25cm}
    \caption{Results for iterations on datasets, accuracy scores are reported as percentages. \dataset{lb} is used for "Test set Acc." since submission scores are not available for all datasets. Marginal and cumulative gains are with respect to validation set. Note that the gain between \dataset{raw} and \dataset{cleaned} is not reported as the validation splits are different, however, it is fixed for the rest. }
    \label{tab:results}
\end{table}

\medskip
\bibliographystyle{unsrtnat}
\bibliography{references}

\begin{thebibliography}{9}
\providecommand{\natexlab}[1]{#1}
\providecommand{\url}[1]{\texttt{#1}}
\expandafter\ifx\csname urlstyle\endcsname\relax
  \providecommand{\doi}[1]{doi: #1}\else
  \providecommand{\doi}{doi: \begingroup \urlstyle{rm}\Url}\fi

\bibitem[{DeepLearning.AI and Landing AI}(2021)]{dccomp2021}
{DeepLearning.AI and Landing AI}.
\newblock Data-centric ai competition, 2021.
\newblock URL \url{https://https-deeplearning-ai.github.io/data-centric-comp/}.

\bibitem[LeCun et~al.(1998)LeCun, Bottou, Bengio, and
  Haffner]{lecun1998gradient}
Yann LeCun, L{\'e}on Bottou, Yoshua Bengio, and Patrick Haffner.
\newblock Gradient-based learning applied to document recognition.
\newblock \emph{Proceedings of the IEEE}, 86\penalty0 (11):\penalty0
  2278--2324, 1998.

\bibitem[He et~al.(2016)He, Zhang, Ren, and Sun]{he2016deep}
Kaiming He, Xiangyu Zhang, Shaoqing Ren, and Jian Sun.
\newblock Deep residual learning for image recognition.
\newblock In \emph{Proceedings of the IEEE conference on computer vision and
  pattern recognition}, pages 770--778, 2016.

\bibitem[Pedregosa et~al.(2011)Pedregosa, Varoquaux, Gramfort, Michel, Thirion,
  Grisel, Blondel, Prettenhofer, Weiss, Dubourg, Vanderplas, Passos,
  Cournapeau, Brucher, Perrot, and Duchesnay]{scikit-learn}
F.~Pedregosa, G.~Varoquaux, A.~Gramfort, V.~Michel, B.~Thirion, O.~Grisel,
  M.~Blondel, P.~Prettenhofer, R.~Weiss, V.~Dubourg, J.~Vanderplas, A.~Passos,
  D.~Cournapeau, M.~Brucher, M.~Perrot, and E.~Duchesnay.
\newblock Scikit-learn: Machine learning in {P}ython.
\newblock \emph{Journal of Machine Learning Research}, 12:\penalty0 2825--2830,
  2011.

\bibitem[Moore and Corso(2020)]{moore2020fiftyone}
B.~E. Moore and J.~J. Corso.
\newblock Fiftyone, 2020.
\newblock URL \url{https://github.com/voxel51/fiftyone}.

\bibitem[Graves(2013)]{graves2013generating}
Alex Graves.
\newblock Generating sequences with recurrent neural networks.
\newblock \emph{arXiv preprint arXiv:1308.0850}, 2013.

\bibitem[Vasquez(2018)]{svasquez-handwriting}
Sean Vasquez.
\newblock {Handwriting Synthesis}, 02 2018.
\newblock URL \url{https://github.com/sjvasquez/handwriting-synthesis}.

\bibitem[Bitton and Papakipos(2021)]{bitton2021augly}
Joanna Bitton and Zoe Papakipos.
\newblock Augly: A data augmentations library for audio, image, text, and
  video.
\newblock \url{https://github.com/facebookresearch/AugLy}, 2021.

\bibitem[{The ImageMagick Development Team}(2021)]{imagemagick}
{The ImageMagick Development Team}.
\newblock Imagemagick, 01 2021.
\newblock URL \url{https://imagemagick.org}.
\newblock version 7.0.10.

\end{thebibliography}

\end{document}